%% file: paper_regular.tex
\documentclass{article}
\usepackage{spconf,amsmath,graphicx}
\usepackage{tipa}
\usepackage{cite}
\usepackage{microtype}
\usepackage{enumitem}
\usepackage{siunitx}
\usepackage{footnote}
\usepackage{tablefootnote}
\usepackage{float}
\usepackage{amssymb}
\usepackage{booktabs}
\usepackage{multirow}
\usepackage{hyperref}

\newcommand{\klcomment}[1]{\textcolor{red}{#1 (KL)}}
\newcommand{\yhcomment}[1]{\textcolor{teal}{#1 (YH)}}
\newcommand{\sscomment}[1]{\textcolor{RedViolet}{#1 (SS)}}
\newcommand{\kledit}[1]{\textcolor{magenta}{#1}}
\newcommand{\yhedit}[1]{\textcolor{blue}{#1}}
\newcommand{\ssedit}[1]{\textcolor{RedOrange}{#1}}

\renewcommand{\klcomment}[1]{}
\renewcommand{\yhcomment}[1]{}
\renewcommand{\sscomment}[1]{}
\renewcommand{\kledit}[1]{{#1}}
\renewcommand{\yhedit}[1]{{#1}}
\renewcommand{\ssedit}[1]{{#1}}

\title{Acoustic Span Embeddings\\ for Multilingual Query-by-Example Search}

\twoauthors{Yushi Hu}{University of Chicago, USA\\\small\texttt{hys98@uchicago.edu}}{Shane Settle, Karen Livescu}{TTI-Chicago, USA\\\small\texttt{\{settle.shane,klivescu\}@ttic.edu}}

\begin{document}
%
\maketitle
\input{abstract}
\input{intro}
\input{related_work}

\input{approach}
\input{setup}
\input{results}

\input{conclusion}

\bibliographystyle{IEEEbib}
\bibliography{strings,refs}

\end{document}

%% file: abstract.tex
\begin{abstract}
Query-by-example (QbE) speech search is the task of matching spoken queries to utterances within a search collection.  In low- or zero-resource settings, QbE search is often addressed with approaches based on dynamic time warping (DTW).  Recent work has found that methods based on acoustic word embeddings (AWEs) can improve both performance and search speed.  However, prior work on AWE-based QbE has primarily focused on English data and with single-word queries.  In this work, we generalize AWE training to spans of words, producing acoustic span embeddings (ASE), and explore the application of ASE to QbE with arbitrary-length queries in multiple unseen languages.  We consider the commonly used setting where we have access to labeled data in other languages (in our case, several low-resource languages) distinct from the unseen test languages.  We evaluate our approach on the QUESST 2015 QbE tasks, finding that multilingual ASE-based search is much faster than DTW-based search and outperforms the best previously published results on this task.
\end{abstract}

\begin{keywords}
query-by-example, acoustic word embeddings, multilingual, low-resource, zero-resource
\end{keywords}

%% file: intro.tex
\vspace{-.1in}
\section{Introduction}
\label{sec:intro}
\vspace{-.1in}
Query-by-example (QbE) speech search is the task of matching spoken queries
to utterances in a search collection~\cite{parada2009query,Mir2015QueryBE,hazen2009query}.  
QbE search often relies on dynamic time warping (DTW) for comparing variable-length audio segments~\cite{hazen2009query,zhang2009unsupervised,
jansen2012indexing,mantena2013speed,leung2016toward}.
DTW is a dynamic programming approach to determine the similarity between two audio segments by finding their best frame-level alignment~\cite{vintsyuk1968speech,sakoe1978dynamic}.
As a result, DTW-based search methods can be very slow (although they can be sped up with approximate nearest neighbor indexing methods~\cite{jansen2012indexing}), and the results depend on the quality of the frame-level representations~\cite{chen2016unsupervised,leung2016toward}. In addition, it can be difficult to  \ssedit{modify DTW-based search to cover both exact and} approximate query matches well; as a result, the best systems on benchmark QbE tasks often fuse many systems together~\cite{leung2016toward,proencca2016segmented}.

An alternative to DTW-based search is to use acoustic word embeddings (AWE)~\cite{levin+etal_asru13,Kamper_16a,chung2016unsupervised,he+etal_iclr2017,kamper2018truly,yuan2018learning,Holzenberger2018} --- vector representations of spoken word segments --- and compare segments using vector distances between their embeddings.  This approach to QbE search, first demonstrated using template-based 
acoustic word embeddings~\cite{levin+etal_icassp15} and later using discriminative neural embeddings~\cite{settle2017query}, can greatly speed up search, even over speed-optimized DTW, as well as improve performance.  Thus far, this prior work on AWE-based QbE has focused on English data and on single-word queries.  Recent work has studied multilingual acoustic word embeddings~\cite{kamper2020multilingual,hu2020multilingual}, but thus far has tested them only on proxy word discrimination tasks.

This work makes two main contributions.  First, we demonstrate how embedding-based QbE can be effectively applied to multiple unseen languages by 
using embeddings learned on languages with available labeled data. Second, we extend the idea of acoustic word embeddings to multi-word spans to better handle \kledit{queries containing an arbitrary numbers of words}.
We evaluate on the QUESST 2015 benchmark, which \ssedit{includes challenging acoustic conditions, multiple low-resource languages, and both exact and approximate match query settings}. 
\kledit{Our approach outperforms all prior work on this benchmark, while also being much faster than DTW-based search.}

%% file: related_work.tex
\begin{figure*}[t]
    \centering
    \includegraphics[width=\linewidth]{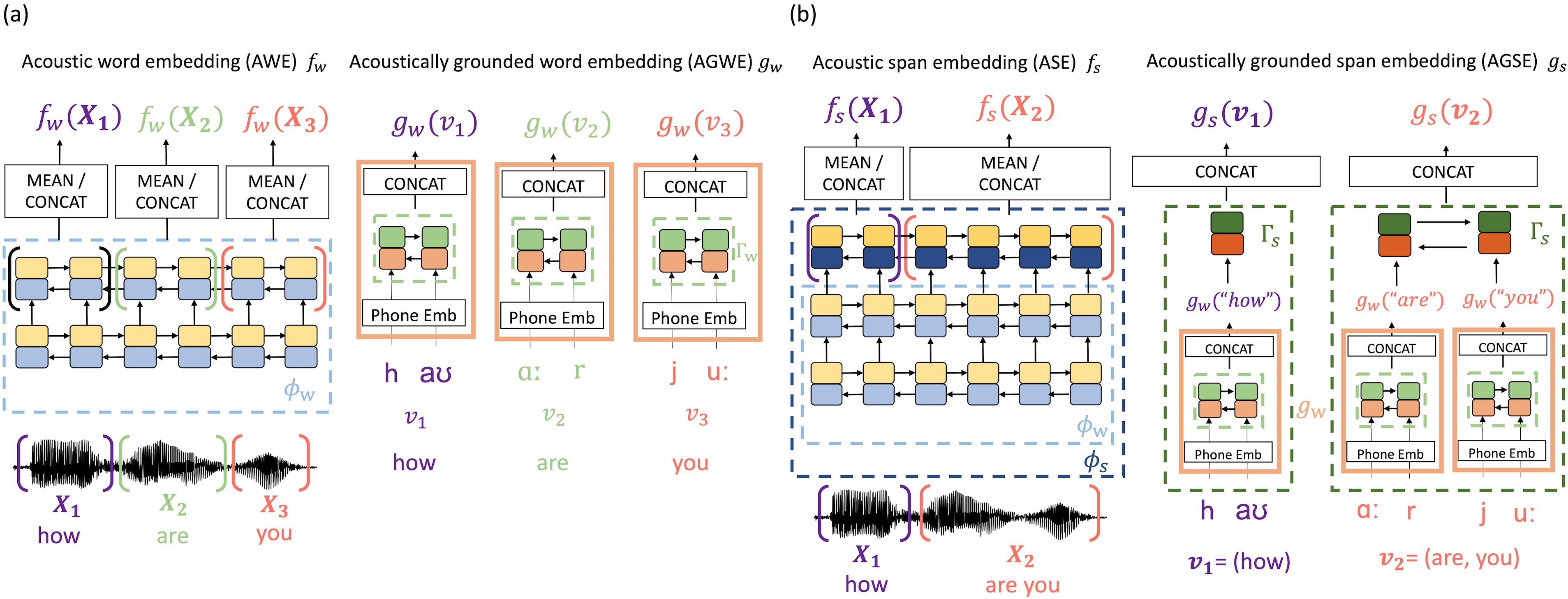}
    \caption{(a) Contextualized acoustic word embedding (AWE) model $f_w$ and acoustically grounded word embedding (AGWE) model $g_w$. (b) Acoustic span embedding (ASE) model $f_s$ and acoustically grounded span embedding (AGWE) model $g_s$. 
    }
    \label{fig:ase}
\end{figure*}

\vspace{-.15in}
\section{RELATED WORK}
\label{sec:related_work}
\vspace{-.1in}


Low-resource QbE methods often use DTW for audio segment comparison~\cite{hazen2009query,zhang2009unsupervised,zhang2011piecewise,mantena2013speed}. 
The frame-level representation is very important in DTW, and a great deal of DTW-based QbE work has focused on learning improved representations.  Typical approaches include using phonetic posteriorgrams~\cite{hazen2009query,zhang2009unsupervised,Lafarga2015ELiRFAM,Lopez-Otero2015GTM} and supervised or unsupervised bottleneck features (BNF)~\cite{proencca2016segmented,DBLP:conf/mediaeval/HouPL0XLXFNXCZS15,leung2016toward,chen2016unsupervised}. 
In addition, many have found
speech activity detection (either energy-based or using a trained phoneme recognizer) to be important for DTW performance and efficiency~\cite{leung2016toward,xu2016approximate,proencca2016segmented}.  Other work has made efficiency improvements through frame-level approximate nearest neighbor search~\cite{jansen2012indexing}.  More challenging query definitions (e.g., \ssedit{approximate} matches) and adverse speech conditions (e.g., noisy or reverberant audio) can challenge DTW-based systems and have motivated the use of system ensembles and data augmentation techniques~\cite{leung2016toward,proencca2016segmented, DBLP:conf/mediaeval/HouPL0XLXFNXCZS15}. Symbolic systems \kledit{based on finite-state transducers 
built from predicted phone sequences} have also been explored to \ssedit{address approximate} matches and improve speed, 
and these can improve performance when fused with more standard DTW systems~\cite{leung2016toward,xu2016approximate}. 
\kledit{As an alternative, the QbE task can be viewed as one of classifying frame-level similarity matrices as matching or non-matching, and learning the classifier in a supervised way
~\cite{ram2020neural}; this allows for end-to-end training, but requires task-specific labeled data.}

\kledit{In contrast,}
embedding-based QbE can be much less complex \kledit{than methods involving frame-level similarities}.
Acoustic word embeddings (AWE) have been explored as an alternative to DTW for comparing variable-length acoustic segments. A number of AWE approaches have been developed, primarily for low- and zero-resource settings, including both supervised~\cite{Kamper_16a,settle2016discriminative,he+etal_iclr2017} and unsupervised~\cite{levin+etal_asru13,audhkhasi2017end,kamper2018truly,Holzenberger2018} approaches. \klcomment{did some shortening}
AWEs are often evaluated on proxy tasks such as word discrimination~\cite{carlin+etal_icassp11}, but 
they have also been applied to downstream tasks including QbE~\cite{levin+etal_icassp15,settle2017query,yuan2018learning} and speech recognition~\cite{BengioHeigol14a,settle2019acoustically}.  For QbE, AWE-based approaches can improve both search performance and querying efficiency~\cite{levin+etal_icassp15,settle2017query,yuan2018learning}.  Recent work has also explored the use of linguistic knowledge in AWE training to better account for approximate matches~\cite{yang2019linguistically}, although this idea has not yet been applied to QbE search.

\kledit{Recent} work has begun to consider the use of multilingual data for learning AWEs.  For example, multilingual BNFs have been used to improve low-resource AWE models on English~\cite{yuan2016bottleneck,yuan2018learning}.
Other work has found that AWEs trained on multilingual data can transfer well to additional zero-resource or low-resource languages, when evaluated on word discrimination~\cite{kamper2020multilingual,hu2020multilingual}.
However, to our knowledge no prior work has explored AWE-based QbE search in a multilingual setting.  In the current work, we extend the multilingual AWEs of~\cite{hu2020multilingual} in several ways for application to multilingual QbE search.


Several benchmark datasets and tasks have been developed for comparing QbE systems, notably the MediaEval series of challenges~\cite{Measures2013,anguera2014query,Mir2015QueryBE}.  In this work we focus on the QUESST QbE task from MediaEval 2015~\cite{Mir2015QueryBE}, which is the most recent QbE task from the MediaEval series and includes a large number of low-resource languages and multiple \ssedit{query settings.} 

%% file: approach.tex

\section{MULTILINGUAL EMBEDDING-BASED QbE}
\label{sec:approach}
\begin{figure}[t]
  \centering
  \includegraphics[width=0.9\linewidth]{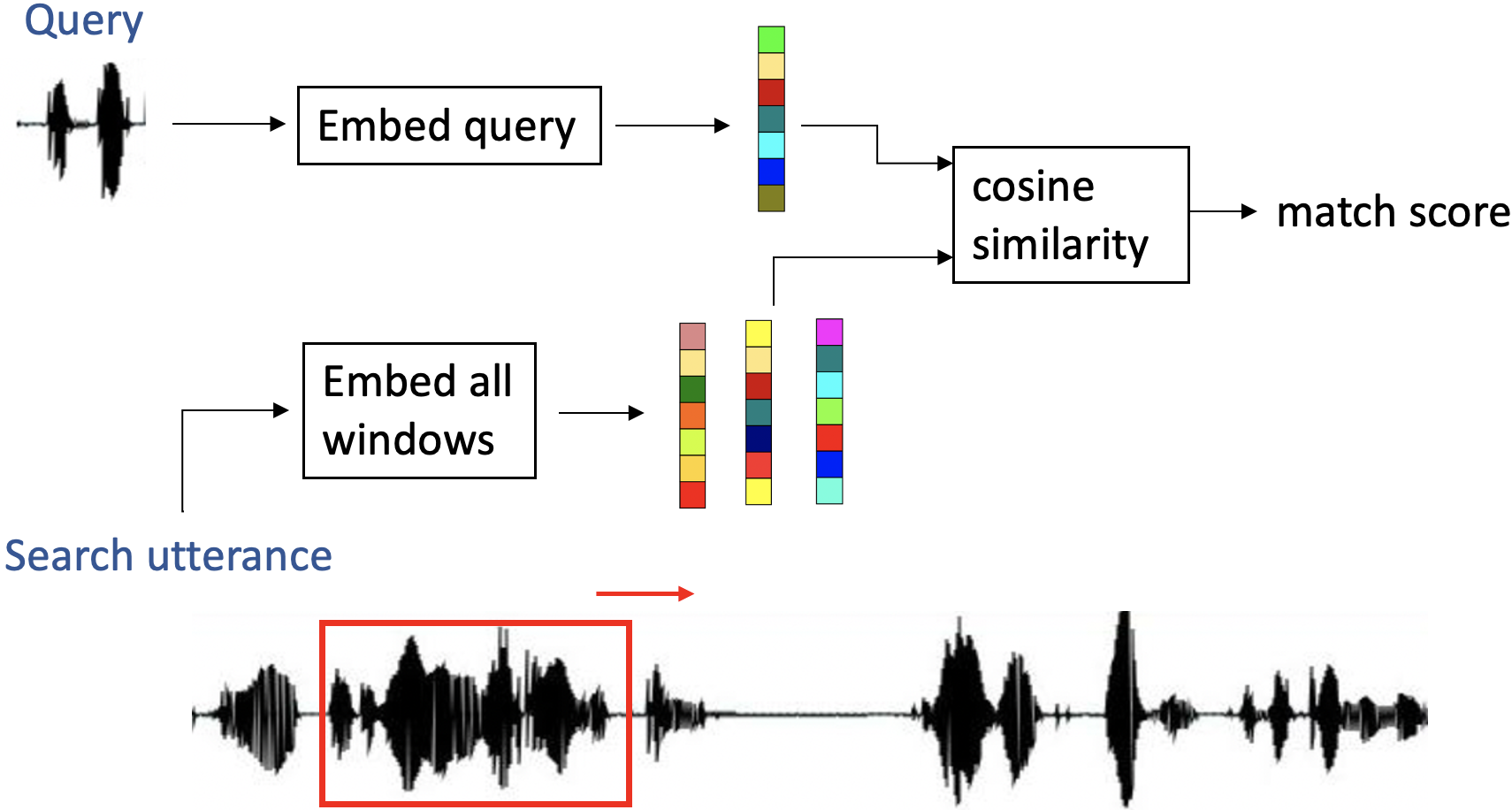}
  \caption{QbE search with acoustic embeddings.  
  }
\label{fig:qbe_system}
\vspace{-0.5cm}
\end{figure}

\vspace{-.1in}
Our QbE 
\kledit{approach, shown in Figure~\ref{fig:qbe_system},} consists of a \kledit{multilingual} \ssedit{embedding} model and a search component. \kledit{The embedding model maps segments of speech---both the query and segments in the search collection---to vectors, and the search component finds matches between embedding vectors. The embedding model is trained 
on data from a set of 
languages, and can then be used for QbE in any language; here we use low-resource languages to train the embedding model, and perform QbE on a disjoint set of unseen languages.} The overall approach is very simple, using no \ssedit{speech} activity detection, fine-tuning for the QbE task, or special-purpose components to accommodate approximate matches; we simply use the trained embeddings out of the box, plugging them into the parameter-free search component.\klcomment{moved this sentence here}

The following subsections describe each component.  For the embedding model, we begin with acoustic word embeddings as has been done in prior work on embedding-based QbE~\cite{levin+etal_icassp15,settle2017query,yuan2018learning}.  
We then extend the 
approach to accommodate spans of multiple words, resulting in {\it acoustic span embeddings}.
Finally, we describe the search component.



\vspace{-.15in}
\subsection{\ssedit{Acoustic word embedding \kledit{(AWE) model}}}
\label{ssec:awe}
\vspace{-.07in}

\kledit{For the acoustic word embedding model,} we take as our starting point an approach that produces the best word discrimination results in prior work, based on jointly learning AWEs and written word embeddings (referred to as acoustically grounded word embeddings (AGWEs)) using a multi-view contrastive loss~\cite{he+etal_iclr2017,hu2020multilingual}, where the AWE is the acoustic ``view" and the AGWE is the written ``view".  We note that we do not use the written embedding model, but we jointly train both models since this approach has produced \kledit{high-quality} AWEs in prior work.  We slightly extend the approach \kledit{in order} to embed word segments in context.  Contextual AWEs improve QbE performance~\cite{yuan2018learning}, simplify extension to multi-word spans (Section~\ref{ssec:ase}), and help 
efficiently embed the search collection.

For AWE learning, we are given a training set of $N$ spoken utterances, $\{{\bf X}^{(n)}\}_{n=1}^N$, where each utterance ${\bf X} \in \mathbb{R}^{T \times D}$ has a corresponding word-level alignment $\mathcal{A}$ used to extract spoken word segments. A word alignment $\mathcal{A} = \{(s_1, e_1, v_1), ..., (s_L, e_L, v_L)\}$ consists of tuples $(s_i, e_i, v_i)$ indicating start frame, end frame, and word label, respectively, for each word in the utterance. \sscomment{removed for convenience $t_i$ as I dont think it's used anymore...} 

Figure~\ref{fig:ase} outlines the structure of the embedding models.  The acoustic-view model $f_w$ consists of an utterance encoder $\Phi_w$ and a pooling function $G$.
The embedding of the $i^\textrm{th}$ segment in ${\bf X}$ is given by the pooled encoder outputs over the frames in the segment:
\vspace{-0.1cm}
\begin{flalign*}
    f({\bf X}_i) &= f({\bf X}, \mathcal{A}_i) = G(\Phi_w({\bf X}), s_i, e_i)
\end{flalign*}
where $\Phi_w({\bf X}) \in \mathbb{R}^{T \times d}$, $\mathcal{A}_i = (s_i, e_i, v_i)$, and $G$ is applied over the window $[s_i, e_i]$.  For the encoder $\Phi_w$, we use a bidirectional recurrent network, so output features within $[s_i, e_i]$ depend on the full utterance context.  We consider two pooling functions $G$, either concatenation of the starting and ending encoder states or a mean over all encoder states in the segment:
\vspace{-0.25cm}
\begin{flalign*}
    &\hspace{-0.2cm}G(\Phi_w({\bf X}), s_i, e_i) = \left[ \Phi_w({\bf X})_{e_i}^{\rightarrow}; \Phi_w({\bf X})_{s_i}^{\leftarrow} \right]\tag{concat}\\
    &\hspace{-0.2cm}G(\Phi_w({\bf X}), s_i, e_i) = \frac{1}{e_i - s_i + 1} \sum_{j={s_i}}^{e_i} \Phi_w({\bf X})_j\tag{mean}
    \vspace{-0.1cm}
\end{flalign*}
where $[{\bf u}; {\bf v}]$ indicates concatenation of vectors ${\bf u}$ and ${\bf v}$ and, for bidirectional network outputs, $\rightarrow$ and $\leftarrow$ index the forward and backward hidden states, respectively.

The written-view model $g_w$ consists of a pronunciation lexicon $\text{Lex}(\cdot)$ mapping words to phones\footnote{\kledit{The approach can be relaxed to not require a pronunciation lexicon, using instead the written character sequence.  Prior work has found an advantage to phones~\cite{hu2020multilingual}, but the performance difference is not large.}}, an encoder $\Gamma_w$, and a pooling function 
(concatenation). 
As in prior work~\cite{hu2020multilingual,he+etal_iclr2017}, 
the embedding of the word $v_i$ is defined as
\vspace{-0.1cm}
\begin{flalign*}
    g(v_i) &= \left[ \Gamma_w({\bf p}_i)_l^{\rightarrow}; \Gamma_w({\bf p}_i)_1^{\leftarrow} \right]
\end{flalign*}
where ${\bf p}_i = \text{Lex}(v_i)$, $l = \text{len}({\bf p}_i)$, and $\Gamma_w({\bf p}_i) \in \mathbb{R}^{l \times d}$.

During training, we minimize the following objective consisting of three contrastive loss terms (updated slightly from~\cite{hu2020multilingual} to improve word discrimination performance):
\vspace{-0.2cm}
\begin{flalign}
    \sum_{n=1}^{N}\sum_{i=1}^{\vert \mathcal{A} \vert} \sum_{obj=0}^{2} \mathcal{L}_{obj}({\bf X}_i^{(n)}, v_i^{(n)})
    \label{eq:mv}
\end{flalign}
Semi-hard negative sampling~\cite{schroff2015facenet} is done for all three \ssedit{terms}:
\begin{flalign*}
    \mathcal{L}_0({\bf X}, v) &= \left[m + d(f({\bf X}), g(v)) - \displaystyle \min_{v' \in \mathcal{V}_0'({\bf X}, v)} d(f({\bf X}), g(v'))\right]_{+}\\
    \mathcal{L}_1({\bf X}, v) &= \left[ m + d(f({\bf X}), g(v)) - \displaystyle \min_{v' \in \mathcal{V}_1'({\bf X}, v)} d(g(v), g(v'))\right]_{+}\\
    \mathcal{L}_2({\bf X}, v) &= \left[ m + d(f({\bf X}), g(v)) - \displaystyle \min_{{\bf X}' \in \mathcal{X}_2'({\bf X}, v)} d(g(v), f({\bf X}'))\right]_{+}
\end{flalign*}
where $f$ is $f_w$, $g$ is $g_w$, $m$ is a margin hyperparameter, $d$ denotes cosine distance $d(a,b) = 1-\frac{a\cdot b}{\Vert a \Vert\Vert b \Vert}$, and
\begin{flalign*}
    \mathcal{V}_0'({\bf X}, v) &:= \{
        v' \vert d(f({\bf X}), g(v')) > d(f({\bf X}), g(v)), v' \in \mathcal{V} / v
    \}\\
        \mathcal{V}_1'({\bf X}, v) &:= \{
        v' \vert d(g(v), g(v')) > d(g(v), f({\bf X})), v' \in \mathcal{V} / v
    \}\\
    \mathcal{X}_2'({\bf X}, v) &:= \{
        {\bf X}' \vert d(g(v), f({\bf X}')) > d(g(v), f({\bf X})), v' \in \mathcal{V} / v 
    \}
\end{flalign*}
\ssedit{where $N$ is the number of utterances and $\mathcal{V}$ is the training vocabulary. In practice, this objective is applied over a mini-batch such that $N$ represents that batch size and $\mathcal{V}$ represents the unique words in that batch. Also, instead of using the single most offending semi-hard negative, we find the top $k$ within the batch and each contrastive loss term in Equation~\ref{eq:mv} is an average over these $k$ negatives. Each mini-batch is from a single language, similar to~\cite{hu2020multilingual}.}

This objective encourages embedding spoken word segments corresponding to the same word label close together and close to their learned label embeddings, while ensuring that segments corresponding to different word labels are mapped farther apart (and nearer to their respective label embeddings).

\vspace{-.1in}
\subsection{\ssedit{Acoustic span embedding} \kledit{(ASE) model}}
\label{ssec:ase}
\vspace{-.05in}
Next, we extend the idea of jointly trained AWE+AGWE 
to jointly trained acoustic span embeddings (ASE) and acoustically grounded span embeddings (AGSE) to better model spans of multiple words that may appear in queries and search utterances.  As before, we train both embedding models jointly, but use only the ASE model for QbE.

The training data for ASE+AGSE learning consists of spans of consecutive word segments, constructed from the word-aligned AWE+AGWE training data by merging neighboring word segments. Specifically, given the training utterance ${\bf X}$ with word-level alignment $\mathcal{A}$, we merge any given pair of consecutive 
word segments $i$ and $j$ with probability $p$, to generate a span-level alignment $\mathcal{A}'$
\begin{flalign*}
    \mathcal{A}' &= \{\ldots, (s_i, e_j, {\bf v}_{i:j}), \ldots\} = \{\ldots, (s_{i'}, e_{i'}, {\bf v}_{i'}), \ldots\}
\end{flalign*}
where $\vert \mathcal{A}' \vert \leq \vert \mathcal{A} \vert$ and $\mathcal{A}'$ consists of tuples $(s_{i'}, e_{i'}, {\bf v}_{i'})$ indicating start frame, end frame, and multi-word label sequence, respectively, for each span in the utterance 
after merging.

As shown in Figure~\ref{fig:ase}, the ASE model $f_s$ has the same form as the AWE function, consisting of an encoder $\Phi_s$ and a pooling function $G$:
\vspace{-0.15cm}
\begin{flalign*}
    f({\bf X}_{i'}) &= f({\bf X}, \mathcal{A}_{i'}) = G(\Phi_s({\bf X}), s_{i'}, e_{i'})
    \vspace{-0.275cm}
\end{flalign*}
where $\Phi_s({\bf X}) \in \mathbb{R}^{T \times d}$, $\mathcal{A}'_{i'} = (s_{i'}, e_{i'}, {\bf v}_{i'})$, and $G$ is applied over the window $[s_{i'}, e_{i'}]$.

The written-view model (the acoustically grounded span embedding model) $g_s$ consists of
\yhedit{an encoder $\Gamma_s$ and another pooling function} (which is, again, concatenation): 
\begin{flalign*}
    g_s({\bf v}_{i'}) &= \left[ \Gamma_s({\bf v}_{i'})_{l'}^{\rightarrow}; \Gamma_s({\bf v}_{i'})_1^{\leftarrow} \right]
    \vspace{-0.2cm}
\end{flalign*}
where ${\bf v}_{i'} = (v_i, \ldots, v_j) ,l' = j - i + 1$, and $\Gamma({\bf w}_{i'}) \in \mathbb{R}^{l' \times d}$.

%
We jointly train $f_s$ and $g_s$ by minimizing a span-level contrastive loss, which has the same form as Equation~\ref{eq:mv}:
\begin{flalign}
    \sum_{n=1}^{N}\sum_{i'=1}^{\vert \mathcal{A}' \vert} \sum_{obj=0}^{2} \mathcal{L}_{obj}({\bf X}_{i'}^{(n)}, {\bf v}_{i'}^{(n)})
    \label{eq:mv2}
    \vspace{-0.25cm}
\end{flalign}
where the 
loss terms are defined as in Section~\ref{ssec:awe} but where $g$ is $g_s$, $f$ is $f_s$, and $\mathcal{V}$ is the set of multi-word spans ${\bf v}$. ASE+AGSEs can be trained from scratch 
using this objective, but in this work we first pre-train word-level models $f_w, g_w$ and use them to initialize the lower layers of $\Phi_s, \Gamma_s$. 
One potential pitfall of a span-level contrastive loss would seem to be that negative examples may be ``too easy". In preliminary experiments, we explored alternative objective functions taking into account not only same and different span pairs but also the degree of difference between them, but these did not outperform the simple contrastive loss described here.

\vspace{-.1in}
\subsection{Embedding-based QbE}
\label{ssec:embqbe}
\vspace{-.05in}
Figure~\ref{fig:qbe_system} depicts our multilingual embedding-based QbE system. \footnote{\url{https://github.com/Yushi-Hu/Query-by-Example}}  Given a pre-trained embedding model (either AWE or ASE), we first build an index of utterances in the search collection by embedding all possible segments that meet certain minimal constraints. Similarly to~\cite{yuan2018learning}, 
we use a simple sliding window algorithm \yhedit{with several window sizes to get possible matching segments. 
}
The 
segment in each analysis window is then mapped to an embedding vector using one of our AWE or ASE embedding models. Contextual AWEs and ASEs help to simplify this process, since each utterance can be forwarded just once through the embedding network ($\phi_w$ or $\phi_s$); the segment embedding for each segment in the utterance is then computed by pooling over the relevant windows.  

At query time, given an audio query, we embed the query with the same pre-trained embedding model, and then compute a detection score 
for each utterance in the search collection via cosine similarity between the corresponding embeddings:
\vspace{-0.2cm}
\begin{equation}
    score({\bf q},{\bf S}) = \underset{{\bf s} \in \Sigma_{\bf q}}{max}{\frac{f({\bf s}) \cdot f({\bf q})}{\Vert f({\bf s}) \Vert_2 \Vert f({\bf q}) \Vert_2}}
\label{eq:score}
\end{equation}
where $f$ is the acoustic-view embedding model (either AWE or ASE),
${\bf q}$ is the spoken query, ${\bf S}$ is the search utterance, and $\Sigma_{\bf q}$ is the set of all windowed segments in $S$ of similar length to ${\bf q}$.  We compute this score for all possible segments in the search collection, but it is possible to speed up the search via approximate nearest neighbor search (as in previous work~\cite{levin+etal_icassp15,settle2017query}).  In this work, we are not concerned with obtaining the fastest QbE system possible, so we simply use exhaustive search; but as we will see in Section~\ref{ssec:time}, this exhaustive embedding-based search is still much faster than DTW-based alternatives.

%% file: setup.tex
\vspace{-.1in}
\section{EXPERIMENTAL SETUP}
\label{sec:setup}
\vspace{-.1in}

\subsection{Embedding \kledit{models}}
\label{ssec:embedding_setup}
\vspace{-.05in}
We train our embedding models \kledit{(Figure~\ref{fig:ase}) on conversational speech} from 12 languages including $22$ hours of English from Switchboard~\cite{godfrey1992switchboard} and $20$-$120$ hours from each of 11 languages\footnote{Cantonese (122 hrs), Assamese (50 hrs), Bengali (51 hrs), Pashto (68 hrs), Turkish (68 hrs), Tagalog (74 hrs), Tamil (56 hrs), Zulu (52 hrs), Lithuanian (33 hrs), Guarani (34 hrs), and Igbo (33 hrs).}
in the IARPA Babel project~\cite{babel_data}. 
We use Kaldi Babel recipe \texttt{s5d}~\cite{povey2011kaldi} to compute acoustic features and extract word alignments for the Babel languages. We use $39$-dimensional acoustic features consisting of $36$-dimensional log-Mel spectra and $3$-dimensional (Kaldi default) pitch features. All other details are the same as~\cite{hu2020multilingual}.
\kledit{In some experiments, we augment the data with SpecAugment~\cite{park2019specaugment}, using one frequency mask ($m_F=1$) with width $F \sim \mathcal{U}\{0, 9\}$ 
and one time mask ($m_T=1$) with width $T \sim \mathcal{U}\{0, \frac{t_{min}}{2}\}$, 
 where $t_{min}$ is the length of the shortest word segment in the utterance such that no word is completely masked. }

For AWE+AGWE training, the acoustic-view model ($f_w$) encoder $\Phi_w$ is a $4$-layer bidirectional gated recurrent unit (BiGRU~\cite{cho2014learning}) network with dropout rate $0.4$, and the pooling function $G$ is 
\kledit{either mean or concatenation}. The written-view model ($g_w$) encoder $\Gamma_w$ is a phone embedding layer followed by a $1$-layer BiGRU.
For ASE+AGSE training, the acoustic-view model ($f_s$) encoder $\Phi_s$ is a $6$-layer BiGRU network with dropout rate $0.4$. The written-view model ($g_s$) encoder $\Gamma_s$ is 
\yhedit{a $1$-layer BiGRU on top of a AGWE submodule that has the same structure as $g_w$.}\klcomment{might need to update wording here, depending on how we resolve the comments about $g_w, \Gamma_s$ in the approach section} Before span-level training, we first train AWE+AGWE models $f_w$ and $g_w$ to be used as initialization. The bottom $4$ layers of $\Phi_s$ are initialized with $\Phi_w$ (from $f_w$) and are kept fixed during training. 
\yhedit{The bottom word-level submodule in $\Gamma_s$ is also initialized with $g_w$}  and not updated during training. All recurrent models use 256 hidden dimensions per direction per layer and generate 512-dimensional embeddings. 

During ASE+AGSE training, to convert word-level alignment $\mathcal{A}$ (length $L$) into span-level alignment $\mathcal{A}'$, we construct the list of word boundaries $\mathcal{B}_{\mathcal{A}} = \{(e_1, s_2), \ldots, (e_{L-1}, s_{L})\}$. Next, we sample the number of word boundaries to remove $r \sim \mathcal{U}(\frac{L-1}{2}, L - 1)$, and then \ssedit{sample $r$ tuples from $\mathcal{B}_{\mathcal{A}}$. For each of the tuples $(e_i, s_{i+1})$, we merge the segments $i$ and $i+1$ such that $\{\ldots, (s_i, e_i, v_i), (s_{i+1}, e_{i+1}, v_{i+1}), \ldots\}$ becomes $\{\ldots, (s_i, e_{i+1}, {\bf v}_{i:i+1}), \ldots\}$, giving us $\mathcal{A}'$.}
\sscomment{okay I think I fixed this now}
\klcomment{I do find this confusing, both because of the difference from the approach section and because I just don't quite follow what is done with the tuples}
The margin $m$ in Equation~\ref{eq:mv} is $0.4$ and the mini-batch size is $\sim30000$ acoustic frames. We start with sampling $k=64$ negatives and gradually reduce to $k=20$. We use Adam~\cite{KingmaBa15a} to perform mini-batch optimization, with initial learning rate 0.0005 and $L2$ weight decay \kledit{parameter} 0.0001. All other settings are the same as in~\cite{hu2020multilingual}. Hyperparameters are tuned using cross-view word discrimination performance on the dev sets as in~\cite{hu2020multilingual}. 

\vspace{-.125in}
\subsection{QbE system}
\label{ssec:qbe_setup}
\vspace{-.05in}
\kledit{We preprocess} each utterance in the search collection, using a sliding window \kledit{approach},
to generate a set of possible segments \kledit{and embed all of the segments using one of our (AWE/ASE) embedding models}. 
\kledit{Specifically, we consider the set of 
possible segments to be 
windows of size $\{12, 15, 18, \ldots, 30, 36, 42, 48, \ldots, 120\}$ frames, with shifts of $5$ frames between windows.}
We embed the query (length $l_q$) using the same AWE/ASE model and \kledit{compute the cosine similarity between the query embedding and} those of all windowed segments 
with length between $\frac{2}{3}l_q$ and $\frac{4}{3}l_q$.

\vspace{-.125in}
\subsection{Evaluation}
\label{ssec:evaluation}
\vspace{-.05in}
We evaluate our models on the Query by Example Search on Speech Task (QUESST) at Mediaeval 2015~\cite{Mir2015QueryBE}. \kledit{The task uses a dataset containing speech in} 6 languages (Albanian, Czech, Mandarin, Portuguese, Romanian, and Slovak). There are about 18 hours of test utterances in the search collection, 445 development queries, and 447 evaluation queries. \yhcomment{The queries are relatively short (5.8 seconds on average). I just delete this sentence}\klcomment{is this the duration of the queries themselves, or of the utterances they are embedded in?} The dataset includes artificially added noise and reverberation with equal amounts of clean, noisy, reverberated, and noisy+reverberated speech.
The task includes three types of \kledit{sub-tasks:} ${\bf T1}$ (exact match), ${\bf T2}$ (allowing word reordering and lexical variations), and ${\bf T3}$ (like ${\bf T2}$, but conversational queries in context).

\vspace{-.1in}
\subsubsection{Evaluation metrics}
\label{ssec:metrics}
\vspace{-.05in}
We use the official \kledit{QUESST 2015 grading scripts}~\cite{Mir2015QueryBE}.
\kledit{The QbE system outputs a score (in our case, a cosine similarity)} for each pair of query ${\bf q}$ and utterance ${\bf S}$. A threshold $\theta$ can be set such that if $score({\bf q}, {\bf S}) > \theta$, then $q$ is considered a hit. The evaluation script varies $\theta$ to compute normalized cross entropy ($C_{nxe}$) and term-weighted value ($TWV$) as defined in~\cite{Measures2013, Mir2015QueryBE}. \ssedit{In QUESST 2015, $C_{nxe}$ is considered the primary metric.} The final results are reported as $min C_{nxe} = \min_\theta C_{nxe}$ and $max TWV = \max_\theta TWV$, that is the best values achieved for the metrics by varying the threshold.\klcomment{removed footnote, for length}


Specifically, the normalized cross entropy $C_{nxe}$ is 
the ratio between the cross entropy of the QbE system output scores \kledit{(measured against the binary ground truth)} and the cross entropy of random scoring.
It ranges between $[0,1]$, and lower $C_{nxe}$ \kledit{values are better}.
The term weighted value $TWV$ is based on hard decisions and is defined as $1- (P_{miss}(\theta) + \beta P_{fa}(\theta))$ where $\theta$ is a threshold, $P_{miss}$ is the miss rate, $P_{fa}$ is the false alarm rate, and $\beta$ is a term weighing the cost of misses vs.~false alarms.
$TMV$ ranges from $-\beta$ to $1$, and higher values are better. 
The official script uses $\beta=12.49$.

%% file: results.tex
\vspace{-.1in}
\section{RESULTS}
\label{sec:results}
\vspace{-.1in}
\subsection{Comparison with prior work}
\label{ssec:comparison_prior}
\vspace{-.05in}


Table~\ref{tab:baselines} compares our work with the top 
QUESST 2015 submissions as well as the best follow-up work~\cite{leung2016toward}. 
\yhedit{Both the AWE and ASE models we use here are 6 layers for a fair comparison.}
Many QUESST systems are trained on 
labelled data (both in- and out-of-domain with respect to the target languages) and use augmentation techniques to limit mismatch with the test data.
\kledit{In addition,} speech activity detection (SAD) and \kledit{system fusion} 
significantly improve performance, and all of the top-performing QUESST systems use both.\sscomment{we probably want to make sure we're consistent with using either SAD and VAD}

\ssedit{Our AWE (mean) model with SpecAugment is competitive with \kledit{some of the top prior $min C_{nxe}$ results}, but there \kledit{is a clear benefit} to span-level embeddings.}
Although our approach is much simpler,
our best single system, ASE (mean) with SpecAugment, 
\klcomment{removed "significantly" -- shouldn't say that in a technical paper unless you perform a significance test} outperforms all prior work on the primary metric $min C_{nxe}$. On the $max TWV$ metric, this system matches all but two of the fusion models in prior work (\cite{DBLP:conf/mediaeval/HouPL0XLXFNXCZS15} and~\cite{leung2016toward}). To compete with these fusion systems, we create our own simple fusion model by summing the scores output by our two ASE (mean) and ASE (concat) systems (both trained with SpecAugment), which gives us an additional performance boost, 
matching $max TWV$ of the best prior fusion models and outperforming them in $min C_{nxe}$ by a wide margin.

\begin{table*}[ht]
  \centering
\small
  \caption{QUESST 2015 performance on dev and eval sets measured by $min C_{nxe}$ and $max TWV$.  
  Training languages are \kledit{separated into} in- and out-of-domain. All SAD systems are based on phone recognizers. \sscomment{I'm not a huge fan of vertical lines, so I removed them, but we can put them back if the consensus is otherwise} \klcomment{numbers in a column should be aligned by place value.  added up and down arrow next to $min C_{nxe}$ and $maxTMV$ as is sometimes done to indicate which is better}
   }
  \begin{tabular}{lccccccll}
    \toprule
    \textbf{Method} & \textbf{systems} & \multicolumn{2}{c}{\textbf{languages}}& \textbf{labeled data}\tablefootnote{``+" denotes \kledit{that} some dataset sizes are unavailable.} & \textbf{SAD} & \textbf{Augmentation} & \multicolumn{2}{c}{$min C_{nxe} \downarrow$ / $max TWV \uparrow$} \\
     &  & in & out & hours & &  & \multicolumn{1}{c}{dev} & \multicolumn{1}{c}{eval} \\
    \midrule
    \textbf{Top prior \kledit{results}}&&&&&&&\\
    $\,\,$ BNF+DTW~\cite{proencca2016segmented} & 36 & 2 & 4 & 384+ &  Yes & noise & 0.778 / 0.234 & 0.787 / 0.206 \\
    $\,\,$ BNF+DTW~\cite{DBLP:conf/mediaeval/HouPL0XLXFNXCZS15} & 66 & 2 & 15 & 643+  &  Yes & noise + reverb & 0.757 / 0.286 & 0.747 / 0.274 \\
    $\,\,$ Exact match fusion~\cite{leung2016toward} & 2 & 0&2 & 423 & Yes & noise + reverb &  0.795 / 0.256 \\
    $\,\,$ Partial match + symbolic~\cite{leung2016toward} & 2 & 0&1 & 260 & Yes & noise + reverb &  0.783 / 0.231 \\
    $\,\,$ Fusion of above two~\cite{leung2016toward} & 4 &0 &2 & 423 &  Yes & noise + reverb & 0.723 / 0.320 \\
    \textbf{Our systems}&&&&&&&&\\
    $\,\,$ AWE (concat) & 1 & 0&12 & 664 &  &  &  0.845 / 0.084 &  \\
    $\,\,$ AWE (mean) & 1 &0 &12 & 664 & & &  0.803 / 0.101 &  \\
    $\,\,$ AWE (mean) & 1 &0 &12 & 664 & & SpecAugment &  0.782 / 0.135 &  \\
    $\,\,$ ASE (concat) & 1 & 0&12 & 664 & & &  0.753 / 0.193 & \\
    $\,\,$ ASE (mean) & 1 & 0&12 & 664 & & &  0.728 / 0.239 & \\
    $\,\,$ ASE (mean) & 1 & 0&12 & 664 & & SpecAugment &  \textbf{0.706 / 0.255} & \textbf{0.692 / 0.246}\\
    $\,\,$ ASE (mean+concat)& 2 &0 &12 & 664 & & SpecAugment & \textbf{0.670 / 0.323} & \textbf{0.658 / 0.298}\\
    \bottomrule
  \end{tabular}
  \vspace{-.2in}
\label{tab:baselines}
\end{table*}




\vspace{-.1in}
\subsection{\kledit{Dependence on query sub-task}}
\label{ssec:query_type}
\begin{figure}[h]
 \centering
 \includegraphics[width=\linewidth]{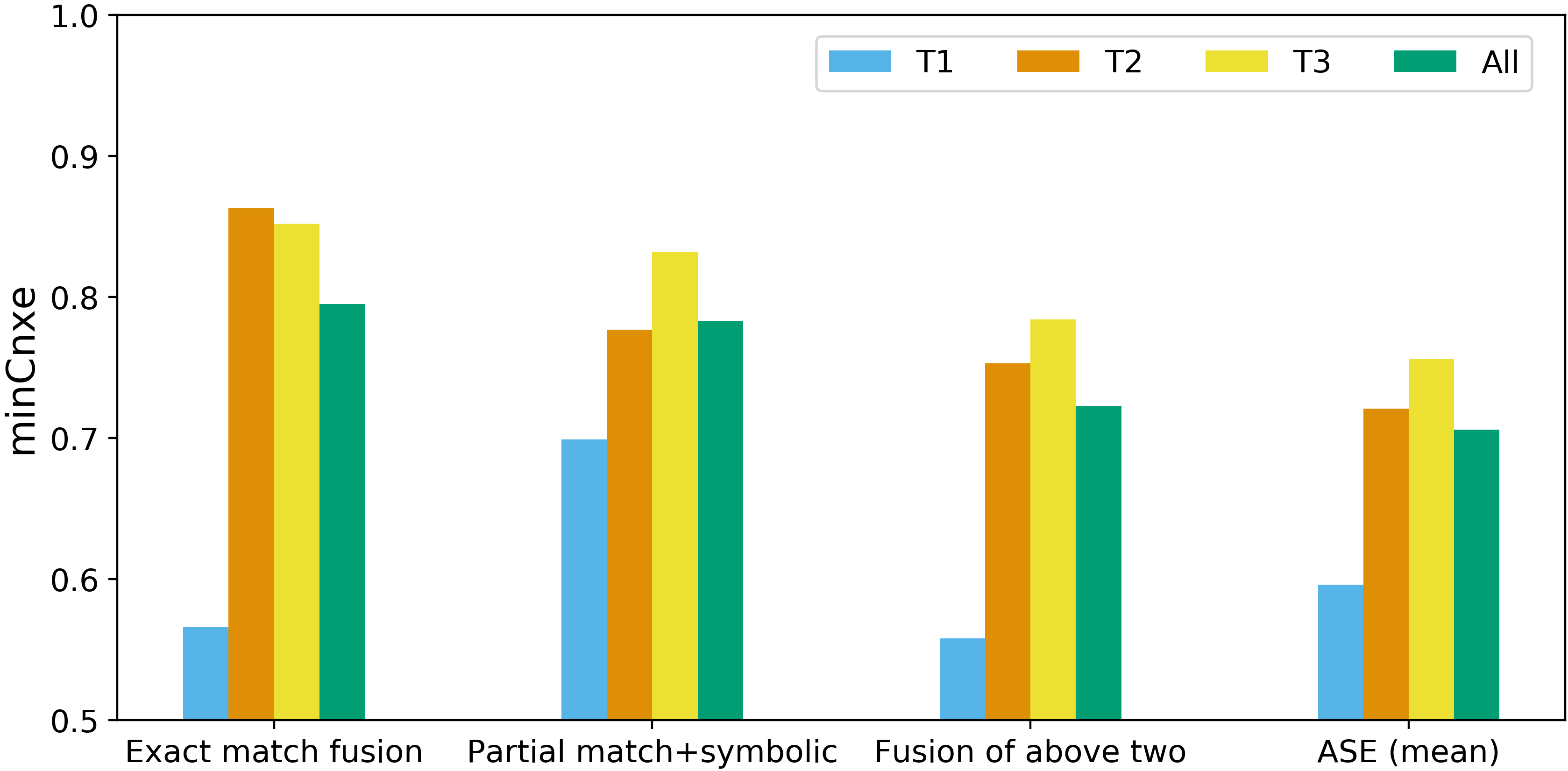}
 \caption{Performance on the three 
 QUESST 2015 \kledit{sub-tasks}, measured by dev set $min C_{nxe}$ ($\downarrow$). 
 }
\label{fig:type_of_queries}
\vspace{-.3cm}
\end{figure}

\vspace{-.05in}
Figure~\ref{fig:type_of_queries} compares $min C_{nxe}$ performance across the QUESST sub-tasks, for both our best single model, ASE (mean) with SpecAugment, and the best models from~\cite{leung2016toward}. We can see that 
the exact DTW system does best in the exact match sub-task (${\bf T1}$) but cannot generalize well to approximate matches (${\bf T2}$ and ${\bf T3}$), while the partial-match system has the opposite behavior. This illustrates the importance of model fusion in DTW-based systems as each individual system can specialize to a particular sub-task.
Meanwhile, our best single system 
is competitive with the others on exact match (${\bf T1}$) and outperforms all of them on approximate match tasks. This \kledit{suggests} that ASE \kledit{models are} better at accommodating 
lexical variations and word re-ordering than DTW-based systems \kledit{without any special-purpose partial-match handling, and without sacrificing too much performance on exact matches}. 
\klcomment{removed the last sentence to save space, and because it was a bit speculative}

\vspace{-.1in}
\subsection{Contribution of segment embeddings} 
\label{ssec:ablation}
\vspace{-.05in}
Our approach differs from prior work in a number of respects, including choice of training data and newer neural network techniques.
One may ask how much of the performance improvements are due to these differences, rather than due to our segment embedding-based approach.  To address this question, we perform a simple comparison on the proxy task of word discrimination, which is often used to compare 
speech representations~\cite{carlin+etal_icassp11,Kamper_16a}. \klcomment{not sure if this is the best wording, or best choice of references}  The task is, given a pair of speech segments each corresponding to a word, to determine whether they correspond to the same word or not.  Specifically we use the task and multilingual development set of~\cite{hu2020multilingual}.  We compare performance on this task 
achieved using 
a vector cosine distance between our acoustic word embeddings, versus using DTW applied to the final layer hidden states
of our network.  Both approaches use the same network trained on the same data, so the only difference is whether we measure distances using segment embeddings computed from the network \ssedit{or} DTW on frame-level network outputs.  On this task, the acoustic word embedding approach has an average precision (AP) of 0.57, whereas the DTW approach has an AP of 0.49.  This comparison demonstrates the improvement due to 
embedding segments as vectors rather than aligning them via DTW.

\vspace{-.15in}
\subsection{Run time}
\label{ssec:time}
\vspace{-.05in}


Table~\ref{tab:run_time} compares the \kledit{average per-query run times} of our implementations of ASE-based and DTW-based QbE search. 
In the DTW-based system, we use a sliding window approach with a window size of $90$ frames and a shift of $10$ frames. This results in each query being compared with $600K$ 
windowed segments, while our ASE-based QbE system on average compares each query with $4M$ windowed segments as described in Section~\ref{ssec:qbe_setup}.
We use 39-dimensional filterbank+pitch features 
and 512-dimensional BiGRU hidden states from the ASE model as inputs to the DTW-based QbE system. 
All systems are tested on a single thread of an Intel Core i7-9700K CPU.
\kledit{As expected, the ASE-based approach is more than two orders of magnitude faster than the DTW-based approaches.}
\kledit{Although the DTW system here is not identical to any of the actual QUESST 2015 systems, this rough comparison gives an idea of the speed difference.  In addition, for emedding-based systems used in a real search setting, approximate nearest neighbor can be used to further speed up the search~\cite{levin+etal_icassp15,settle2017query}}.
\vspace{-0.4cm}

\begin{table}[H]
  \centering
  \small
  \caption{\kledit{Run times on the} QUESST 2015 development set. 
}
  \begin{tabular}{lcr}
    \toprule
     \multirow{2}{*}{\textbf{Method}} & \# of comparisons & Run-time \\ 
     & (per query) & (s / query) \\
    \midrule
    DTW on Filterbank features & 600K & 486\\
    DTW on ASE hidden states  & 600K & 847\\
    \textbf{ASE-based QbE} & \hspace{.1in} 4M & \textbf{5}\\
    \bottomrule
  \end{tabular}
  \vspace{-.2in}
\label{tab:run_time}
\end{table}

%% file: conclusion.tex
\vspace{-.1in}
\section{CONCLUSION}
\label{sec:conclusion}
\vspace{-.05in}

\kledit{We have presented a simple embedding-based approach for multilingual query-by-example search that outperforms prior work on the QUESST 2015 QbE task, while also being much more efficient.}
We verify that multilingual acoustic word embedding (AWE) models can be effective for query-by-example search on unseen target languages, and we further improve performance by extending the approach to multi-word spans using acoustic span embeddings (ASE). 
Our embeddings are jointly trained with embeddings of the corresponding written words, but for QbE we ignore the written embeddings.  In future work it will be interesting to use both of the jointly learned models to search by either spoken or written query.



\klcomment{reminder to fix reference formatting -- capitalization, standardizing conference names etc.}